\documentclass[letterpaper, 10 pt, conference]{ieeeconf}  
\IEEEoverridecommandlockouts    
\usepackage{comment}
\usepackage{cite}
\newcommand{\eat}[1]{}
\usepackage{booktabs}
\usepackage{color}
\usepackage{xcolor, soul}
\usepackage [autostyle, english = american]{csquotes}
\MakeOuterQuote{"}
\usepackage{xcolor}
\let\labelindent\relax 
\usepackage{enumitem}
\usepackage{threeparttable}

\usepackage{multicol}
\usepackage{multirow}

\usepackage{mathtools}
    
\usepackage{amsmath}
\usepackage{amstext}
\usepackage{amssymb}
\usepackage{amsfonts}
\usepackage{float}

\usepackage{amsthm}  
\usepackage[normalem]{ulem} 
\usepackage{subfig}
\usepackage{caption}
\usepackage{todonotes}

\makeatletter

\usepackage{color,soul} 
 
\newcommand{\Rmnum}[1]{\expandafter\@slowromancap\romannumeral #1@}
\usepackage{savesym}

\usepackage{algorithm}
\usepackage{algorithmicx} 
\usepackage{algpseudocode} %
\savesymbol{AND}
\usepackage[group-separator={,},group-minimum-digits={3}]{siunitx}

\usepackage{graphicx} 
\usepackage{epsfig} 

\usepackage{times} 
\usepackage{amsmath} 
\usepackage{amssymb}  
\usepackage{comment}
\makeatletter
\let\NAT@parse\undefined
\makeatother
\usepackage{hyperref}
\hypersetup{
   colorlinks=true,
    linkcolor= blue,
    allcolors=blue,
    citecolor = blue,
    filecolor=black,      
    urlcolor=blue,
    }
\usepackage{mathrsfs}

\usepackage[symbol]{footmisc}

\makeatletter
\renewcommand\@makefntext[1]{%
\setlength\parindent{1em}%
\noindent
\mbox{\@thefnmark}{#1}}
\makeatother

\begin{document}

\title{\LARGE \bf
MR-LDM - The Merge-Reactive Longitudinal Decision Model:  

Game Theoretic Human Decision Modeling for  Interactive Sim Agents}
\author{ Dustin~Holley,
Jovin~D'sa$^{*}$,
Hossein~Nourkhiz Mahjoub,
Gibran~Ali
\thanks{\hangindent=0.5cm D. Holley is with the Global Center for Automotive Performance Simulation}
\thanks{J. D'sa, H. N. Mahjoub are with Honda Research Institute, USA Inc.}
\thanks{G. Ali is with the Division of Data and Analytics, Virginia Tech Transportation Institute}
\thanks{$^{*}$Corresponding author: {\tt jovin\_dsa@honda-ri.com}}
}

\maketitle
\thispagestyle{empty}
\pagestyle{empty}

\begin{abstract} 
Enhancing simulation environments to replicate real-world driver behavior, i.e., more humanlike sim agents, is essential for developing autonomous vehicle technology. In the context of highway merging, previous works have studied the operational-level yielding dynamics of lag vehicles in response to a merging car at highway on-ramps.
Other works focusing on tactical decision modeling generally consider limited action sets or utilize payoff functions with large parameter sets and limited payoff bounds. 
In this work, we aim to improve the simulation of the highway merge
scenario by targeting a 
game theoretic model for tactical decision-making with improved payoff functions and lag actions.
We couple this with an underlying dynamics model to have a unified decision and dynamics model that can capture merging interactions and simulate more realistic interactions in an explainable and interpretable fashion. The proposed model demonstrated good reproducibility of complex interactions when validated on a real-world dataset. The model was finally integrated into a high-fidelity simulation environment and confirmed to have adequate computation time efficiency for use in large-scale simulations to support autonomous vehicle development.
\end{abstract}
\section{Introduction}

Simulation-based evaluation has become an indispensable tool in the development and testing of Intelligent Transportation Systems (ITS), offering a safe and controllable environment for replicating complex real-world interactions. To better assess system performance in high-interaction scenarios such as highway merging, it is beneficial for simulations to be both tunable and representative of real-world traffic behavior.

\begin{figure}
\vspace{2mm}
    \centering
    \includegraphics[width=.45\textwidth,keepaspectratio]{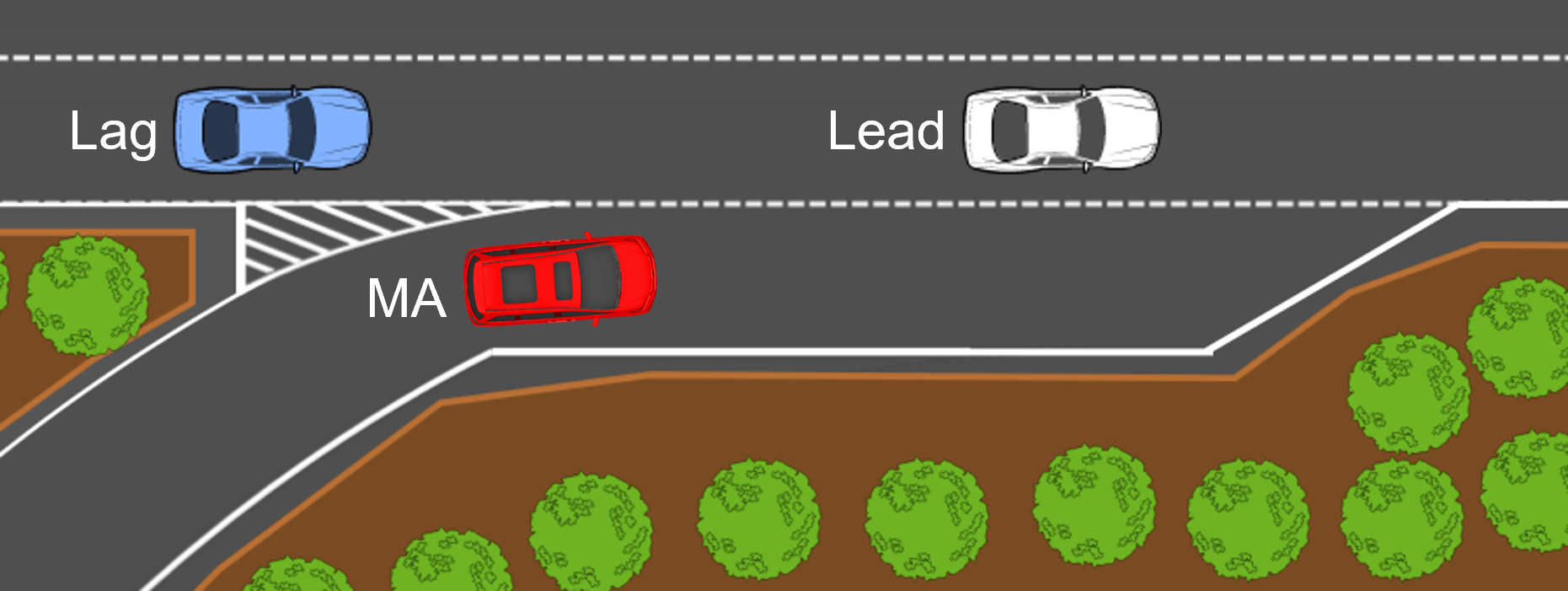}
    \caption{Actor naming conventions.}
    \label{fig:actorlayout}
\end{figure}

In this work, we focus on modeling the behavior of critical traffic actors to support the evaluation of merging trajectory planners in controllable and realistic simulation settings. A key component of this effort involves accurately modeling the behavior of surrounding traffic actors—specifically, the lag actor, i.e., the vehicle in the main lane directly behind the merging vehicle (fig. \ref{fig:actorlayout}). Capturing the behavior of this actor is critical for understanding how merging decisions influence and are influenced by surrounding traffic.

Our previous work introduced the Merge-Reactive Intelligent Driver Model (MR-IDM) \cite{MRIDM}, an extension of the continuous car-following IDM framework \cite{IDM} that reacts to merging vehicles by adjusting gap targets. However, operating at the operational level, MR-IDM’s behavior was implicitly governed by its parameters, limiting its ability to represent distinct decision-level strategies. Likewise, our 
mBRGT-D model \cite{BRGTD}, while explicitly making tactical decisions, focused on lane-change behavior and did not address longitudinal decisions in merging contexts. In this work, we address this gap by developing a new model for the lag actor that explicitly generates discrete, decision-level behaviors in a tunable and interpretable manner. This functionality is essential for systematically testing a wide range of merging planner designs under controlled and varied interaction conditions. 

The proposed model is based on a game-theoretic framework that selects high-level longitudinal actions—such as yielding, maintaining speed, or assertively advancing—which are then executed using MR-IDM dynamics.
It draws conceptual inspiration from \cite{RGLC}, hereafter referred to as the Repeated Game Lane Changing (RGLC) model. While the RGLC model provides a useful starting point, it lacks the flexibility and expressiveness needed to simulate diverse lag driver behaviors. Our model extends the contributions of RGLC by 
expanding the action set of the lag actor, introducing an alternative payoff function and safety metric for more control over actor incentives, and incorporating bounded rationality.

By enhancing the behavioral realism and controllability of traffic actor models, this work contributes to the development of more robust, comprehensive testing environments for intelligent and autonomous driving systems. In particular, it supports the structured evaluation of merging trajectory planners in interactive traffic scenarios where other drivers’ decisions significantly affect outcomes.

\subsection{Related Work}




Understanding and modeling the decision-making behavior of traffic actors—especially in dynamic interactions like highway merging—has been a central theme in ITS research. Mainstream approaches to modeling vehicle interactions include rule-based systems and machine learning (ML) methods. While rule-based models are computationally efficient and interpretable, they often oversimplify behavior and lack generalizability. On the other hand, ML models can capture rich patterns but tend to be data-hungry, computationally intensive, and opaque in terms of decision logic, making them less ideal for real-time, controllable simulation environments.

In contrast, game-theoretic models offer a compelling middle ground. They encode human decision-making structures while remaining interpretable and computationally viable. This has led to a growing body of research on using game theory for modeling merging and lane-changing behavior. Most existing game-theoretic models cast the interaction between merging and main-lane vehicles as a two-player, non-cooperative, non-zero-sum game, where each actor selects strategies to maximize utility (or minimize cost) based on defined payoff functions.

Several studies have developed payoff formulations around goals such as safety, space, comfort, speed, efficiency, and fuel consumption. Metrics commonly used include relative distance, time-to-collision (TTC), time headway, acceleration, and distance to end of ramp \cite{chen2023,RGLC,ALI2019220,yoo_stackelberg_2020}. However, challenges persist in balancing payoff function complexity and interpretability, especially when combining heterogeneous metrics without introducing trivial equilibrium conditions. While models like \cite{chen2023} use a summation of metrics to derive the actor payoffs, normalized payoff functions like the hyperbolic tangent \cite{RGLC} can provide bounds to produce more stability.

Game solutions in the literature include pure and mixed Nash equilibrium, such as in \cite{RGLC}, and Stackelberg formulations, like \cite{ji_lane-merging_2021}. Since Nash equilibria assume perfectly rational actors, a popular choice for modeling realistic traffic interactions is the introduction of bounded rationality through Quantal Response Equilibrium (QRE) \cite{chen2023}, which incorporates probabilistic behavior aligned with human decision uncertainty. Stackelberg games model hierarchical decision-making, often assigning the merging actor as the lead decision-maker. However, such formulations may not capture the simultaneity of real-world interactions. In simulation, this can have the effect of neglecting safety-critical situations caused by conflicting decisions between the traffic and merging actor.

While many games are played at a single decision point (such as the point when the merger reaches the ramp soft nose), models like \cite{RGLC} utilize a repeated game, where actors continuously update decisions over time based on ongoing negotiation. This allows traffic to continue to adapt to a merging vehicle over the entire duration of the merge.
Models that build on this with cumulative payoffs \cite{RGLC}, or time-evolving rationality or politeness \cite{ji_lane-merging_2021} better reflect humanlike adaptive behavior.

Commonly observed longitudinal actions of the lag actor include yielding, blocking, accelerating, and doing nothing. With the exception of some models like \cite{BRGTD}, rarely are all of these behaviors included in a single model's lag action set. For example, the RGLC's lag is limited to \emph{yield} and \emph{block}. This places a limitation on the number of real-world situations the model can produce in a simulation environment.

Building on these observations from the literature, the main contributions of our proposed game-theoretic model are presented in the following section.





\subsection{Contributions}

\begin{itemize}
    \item \textbf{Explicit Longitudinal Decision-Modeling:} We introduce a novel game theory-based Merge-Reactive Longitudinal Decision Model (MR-LDM) that explicitly models four strategies: yield behind, yield ahead, block, and do nothing. Existing  methods either implicitly model degree of yielding or model only limited reactions such as whether to yield or not, or to yield or block.

    \item \textbf{Tunable Behavior using a Custom Payoff Function:} 
    We introduce a modified hyperbolic tangent function (Soboleva tangent) that provides bounded, interpretable, and tunable payoffs, addressing inhomogeneous payoff scales in earlier works. This allows smooth transitions between behavior incentives and better captures humanlike nuance (e.g., aggressive vs. yielding behavior). This also aids in minimizing the size of the parameter set.

    \item \textbf{Novel Input Metric – Predictive Time Headway (PTH):} This metric combines the benefits of TTC and time headway and was shown to offer more realism in behavior execution.

    \item \textbf{Stochasticity via Bounded Rationality (QRE):} We utilized bounded rationality through QRE, enabling probabilistic behavior and capturing human decision inconsistencies.

    \item \textbf{Real-Time Execution:} We show a structured way to integrate longitudinal decision modeling with a downstream dynamics execution that can be effectively used to generate more realistic merging reactions.
\end{itemize}
We use bi-level optimization to validate the model using real-world datasets and show that such interactive behavior execution can be done at a high frequency that is suitable for real-time simulation, even when integrated into a high-fidelity simulator like IPG CarMaker.

The rest of the paper is structured as follows: In section \ref{sec:mr-ldm}, we introduce our proposed model and its different design aspects and payoff design. We then introduce our datasets and data pre-processing steps, along with the model calibration in section \ref{sec:dataset}. We then discuss our behavior execution procedure and simulation integration in section \ref{sec:beh-ex}, followed by concluding remarks.

\section{Methodology} \label{sec:mr-ldm}

\subsection{Behavior Design and Action Space}
In real-world merging scenarios, lag vehicles exhibit a range of behaviors influenced by factors such as driver preferences, traffic density, perceived urgency, etc. While the influencing factors are many, we can reduce the final observable actions into a smaller discrete set. Based on our observations from the HOMER dataset \cite{BRGTD}, we explicitly defined four discrete actions for the lag vehicle in the MR-LDM model:

\begin{itemize}
    \item \textbf{Yield Behind}: The lag vehicle decelerates to allow the merging vehicle to enter ahead.
    \item \textbf{Yield Ahead}: The lag vehicle accelerates to allow the merging vehicle to enter behind, often used when the merger is still relatively far behind or sometimes when adjacent to it.
    \item \textbf{Block}: The lag vehicle accelerates or maintains speed to prevent the merging vehicle from entering the lane, representing a more adversarial driving style.
    \item \textbf{Do Nothing}: The lag vehicle makes no strategic adjustment, maintaining its current speed and spacing.
\end{itemize}

This action set captures both cooperative and adversarial behaviors commonly observed in naturalistic driving. The \textit{block} behavior models more aggressive responses, while \textit{yield ahead} enables proactive cooperation by creating space behind, allowing the car to not slow down for the merger, often seen in the real world. The \textit{do nothing} action reflects hesitation or neutrality when no strong incentives are present. 

\subsection{Game-Theoretic Formulation}
We model the interaction between the merging vehicle (also referred to as ``merging actor'' or \textbf{MA}) and the lag vehicle (\textbf{Lag}), 
as a two-player, non-cooperative, non-zero-sum, repeated game with incomplete information. This formulation allows each driver to repeatedly assess and update their actions over multiple discrete decision points, reflecting realistic driving behavior.
%
%
At each decision step, MA and Lag simultaneously select actions based on their expected payoffs, current road conditions, and previously observed actions. 
Table~\ref{tab:normalForm} presents the normal-form representation of the game, where \textit{P} and \textit{Q} represent MA’s and Lag’s payoffs, respectively, for each pair of actions.  

\begin{table}[htbp]
    \centering
    \caption{Normal form representation of MR-LDM game.}
    \resizebox{\columnwidth}{!}{%
    \begin{tabular}{lcccc}
        \toprule
        & \multicolumn{4}{c}{Lag} \\
        \cmidrule(lr){2-5}
        MA & Yield Behind & Yield Ahead & Block & Do Nothing \\
        \midrule
        Change Lanes & \(P_{11}, Q_{11}\) & \(P_{12}, Q_{12}\) & \(P_{13}, Q_{13}\) & \(P_{14}, Q_{14}\) \\
        Keep Straight & \(P_{21}, Q_{21}\) & \(P_{22}, Q_{22}\) & \(P_{23}, Q_{23}\) & \(P_{24}, Q_{24}\) \\
        \bottomrule
    \end{tabular}}
    \label{tab:normalForm}
\end{table}

\subsection{Soboleva Tangent for Payoffs}
\begin{figure}
\vspace{2mm}
    \centering
    \includegraphics[width=\columnwidth,keepaspectratio]{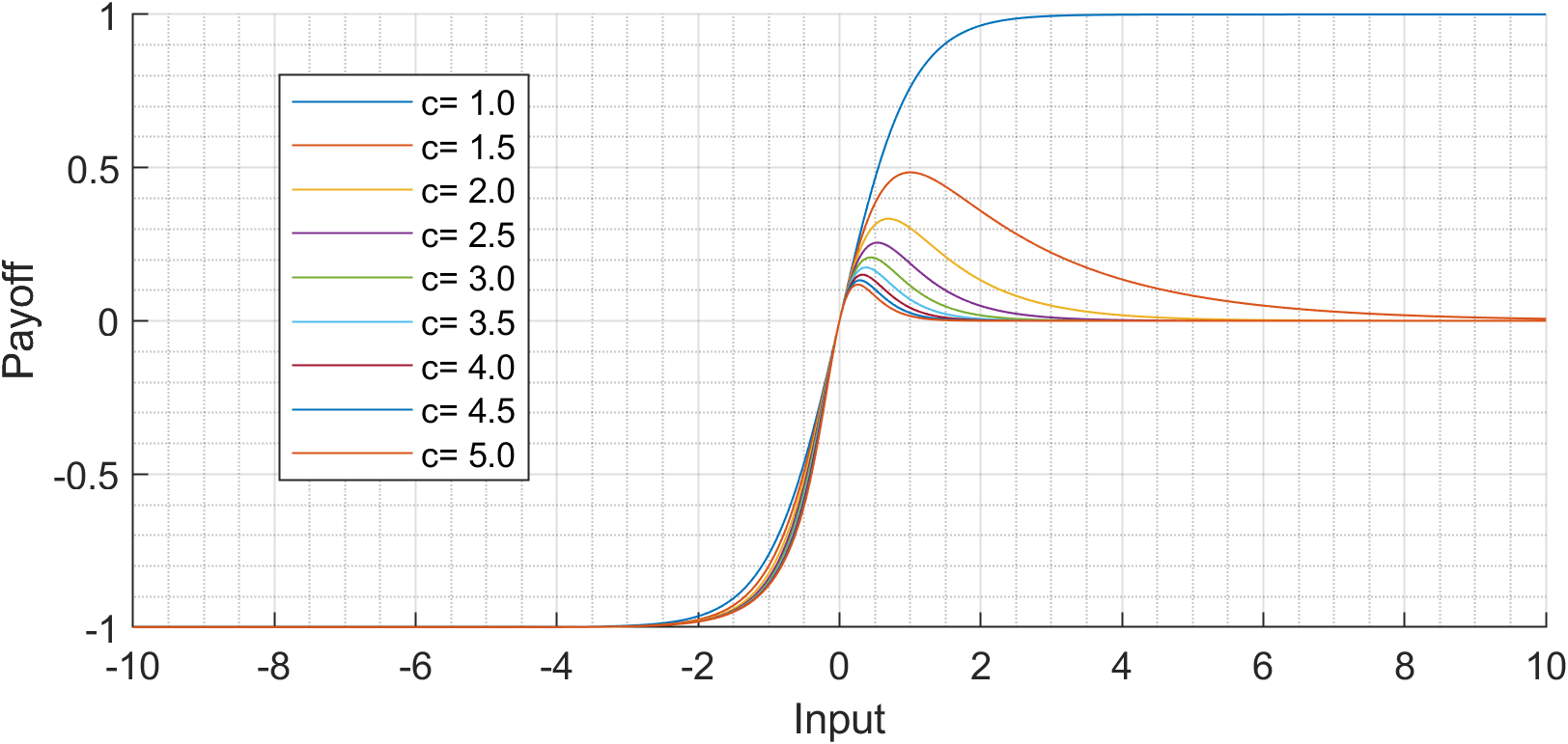}
    \caption{Visualization of the \textit{usmht} function for $a$=$b$=$d$=1 and varying $c$.}
    \label{fig:smhtWithVaryingC}
\end{figure}

Traditional game-theoretic models typically define driver payoffs using linear combinations of metrics such as relative distance, speed, and TTC. However, these linear formulations often produce unbounded and unrealistic utility values under extreme conditions. To address this limitation, we introduce the \textit{updated Soboleva modified hyperbolic tangent (usmht)} function (a modification of the Soboleva modified hyperbolic tangent \cite{smht1,smht2,smht3} shown in Figure \ref{fig:smhtWithVaryingC}), ensuring bounded, interpretable, and smoothly transitioning payoff values.

The \textit{usmht} function is formally defined as:

\[
\text{usmht}(x, c, d, r) = \frac{e^{\,rx + \text{shift}(x, c, d)} - e^{-\,(rx + \text{shift}(x, c, d))}}{e^{\,c(rx + \text{shift}(x, c, d))} + e^{-\,d(rx + \text{shift}(x, c, d))}}
\]

with the shift function defined as:

\[
\text{shift}(x, c, d) = \arg\max_{x}\left(\frac{e^{x} - e^{-x}}{e^{c x} + e^{-d x}}\right)
\]

In our formulation, parameters were selected as follows:
\begin{itemize}
    \item \( c > 1 \) controls the curvature and ensures payoffs taper smoothly at extreme values.
    \item \( d = 1 \) maintains symmetric behavior around the chosen shift point.
    \item \( r \) indicates directionality, with \( r = +1 \) for yield-behind and block behaviors, and \( r = -1 \) for yield-ahead behavior.
\end{itemize}

\begin{figure}
    \centering
    \includegraphics[width=\columnwidth,keepaspectratio]{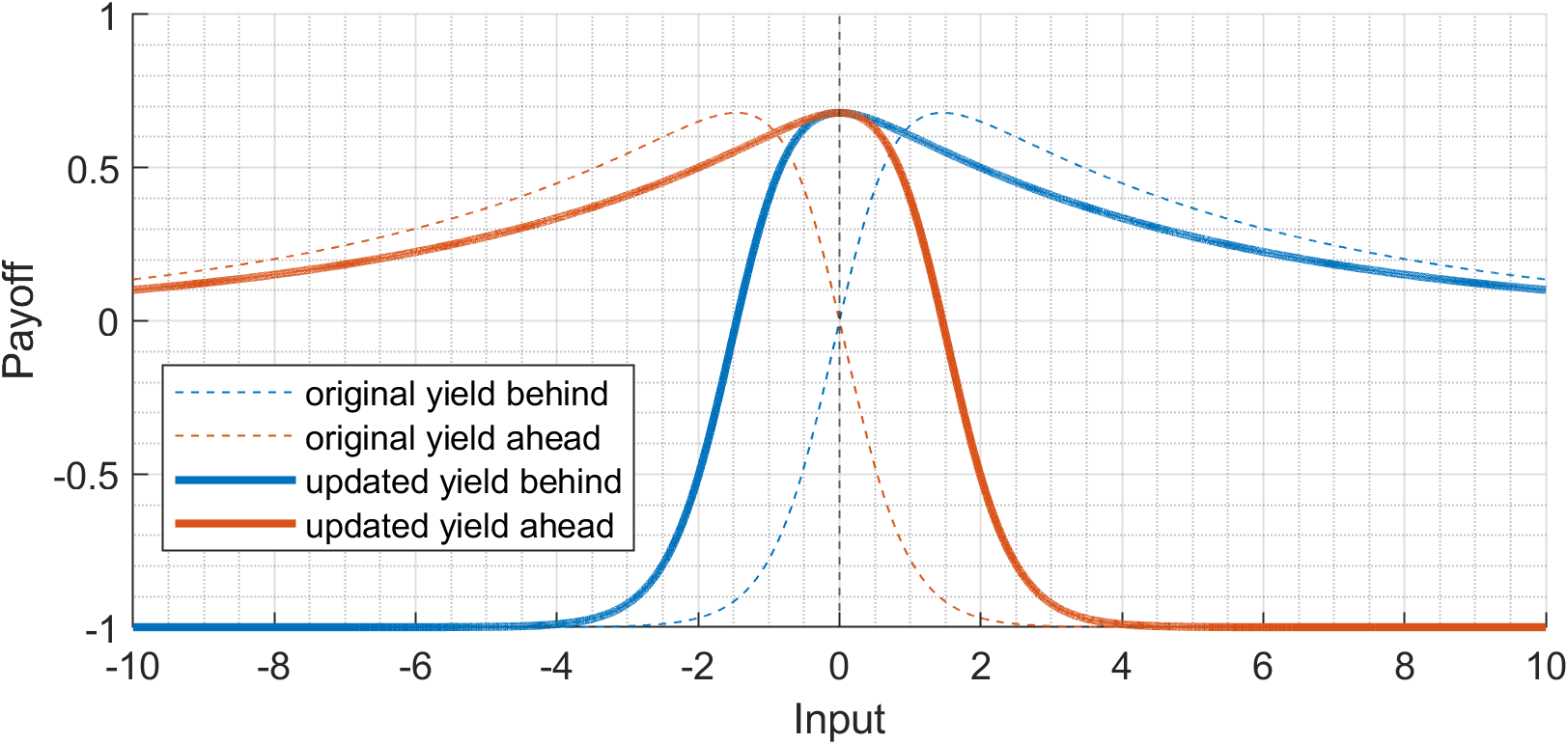}
    \caption{Updated Soboleva modified hyperbolic tangent.}
    \label{fig:usmhtWithShift}
\end{figure}

This modified \textit{usmht} function improves upon the standard hyperbolic tangent (tanh) by allowing asymmetric and adjustable curvature, enabling more precise modeling of human driver preferences. The shape and saturation characteristics of \textit{usmht} facilitate clear differentiation between strong and weak incentives, essential for realistic decision-making models. Figure \ref{fig:usmhtWithShift} visualizes the \textit{usmht} function with the peak shifted to provide a smoother transition between yielding behaviors. By adopting the \textit{usmht} function, MR-LDM achieves greater realism and interpretability, significantly enhancing its ability to model nuanced and humanlike driving behaviors in merging scenarios.

\subsection{Predictive Time Headway (\textit{PTH})}

Traditional metrics such as relative distance or TTC individually fail to fully capture realistic driver decision-making, as they often lead to contradictory or counterintuitive behaviors during merging interactions. For instance, when the MA passes the lag vehicle, the time headway shifts from negative to positive values, while TTC becomes negative, creating conflicting signals that may incorrectly trigger aggressive behavior (e.g., unnecessary acceleration) in conventional models. To address these limitations, we introduce \textit{PTH}, a forward-looking metric that integrates both spatial and dynamic considerations. \textit{PTH} is defined as the predicted time headway from the lag vehicle (Lag) to the MA after a prediction horizon \(\tau\):

\begin{equation}
    PTH^{\mathrm{MA}}_{\mathrm{Lag}}(t) = \frac{\Delta{x}(t+\tau)}{v_{Lag}(t)} = \frac{\Delta{x}(t) + \tau\Delta{v}(t)}{v_{Lag}(t)}
\end{equation}
where $\Delta{x}(t)$ is the relative longitudinal distance from the lag to the merger at time $t$, $\Delta{v}(t)$ is the relative speed, and $v_{Lag}(t)$ is the lag vehicle's speed, utilizing a constant velocity assumption for the vehicles' speeds during the \(\tau\) interval.

Unlike traditional metrics, \textit{PTH} explicitly incorporates anticipated future positions of both vehicles, enabling more intuitive and robust payoff evaluations. This predictive capability allows MR-LDM to model proactive behaviors such as preemptive yielding or aggressive blocking.  By resolving conflicts common in traditional metrics during rapid relative position changes, \textit{PTH} supports a more nuanced and adaptive representation of driver behavior within dynamic merging interactions.

\subsection{Payoff Formulation}

Each action available to the Lag and MA is associated with a payoff that quantifies the expected utility of selecting that behavior. These payoffs are computed using the \textit{usmht} applied to the \textit{PTH}, ensuring smooth and bounded incentive values.

To simplify notation, we define:
\[
\Psi_{\text{Lag,MA}} = \frac{\text{PTH}_{\text{Lag}}^{\text{MA}}}{s_{\text{lat}} s_{\text{ramp}}}, \quad
\Psi_{\text{Lag,Lead}} = \frac{\text{PTH}_{\text{Lag}}^{\text{Lead}}}{s_{\text{lat}} s_{\text{ramp}}}
\]
\[
\Psi_{\text{MA,Lead}} = \frac{\text{PTH}_{\text{MA}}^{\text{Lead}}}{s_{\text{lat}} s_{\text{ramp}}}, \quad
\Psi_{\text{MA,Lag}} = \frac{\text{PTH}_{\text{MA}}^{\text{Lag}}}{s_{\text{lat}} s_{\text{ramp}}}
\]

where each \(\Psi\) represents a scaled predictive time headway, normalized by lateral and ramp-end urgency factors.

\subsubsection{Lag Vehicle Payoffs}

The lag vehicle selects from four actions: Yield Behind (YB), Yield Ahead (YA), Block (Bk), and Do Nothing (DN). The payoffs are defined as:

\begin{align}
Q^*_{\text{YB}} &= \text{usmht}\left( \Psi_{\text{Lag,MA}}, \phi_1, 1, 1 \right) \label{eq:qyb}
\end{align}

\begin{equation}
\begin{split}
Q^*_{\text{YA}} = \text{usmht}\left( \Psi_{\text{Lag,MA}}, \phi_2, 1, -1 \right) \\
\quad - \text{usmht}\left( \Psi_{\text{Lag,Lead}}, \phi_3, 1, 1 \right)
\end{split}
\label{eq:qya}
\end{equation}

\begin{align}
Q^*_{\text{DN}} &= \phi_6 \label{eq:qdn} \\
Q^*_{\text{Bk}} &= \text{usmht}\left( \Psi_{\text{Lag,MA}}, \phi_7, 1, 1 \right) \label{eq:qblock}
\end{align}

where:
\begin{itemize}[leftmargin=1.5em]
    \item \(Q^*_{\text{YB}}\): Payoff for yielding behind by decelerating.
    \item \(Q^*_{\text{YA}}\): Payoff for yielding ahead, adjusted by influence from the leader.
    \item \(Q^*_{\text{DN}}\): Constant reward for taking no strategic action.
    \item \(Q^*_{\text{Bk}}\): Payoff for blocking the merger.
\end{itemize}

\subsubsection{Ramp End Influence Terms}

We introduced scaling terms to dynamically adjust payoffs based on lateral proximity and ramp position:

\begin{align}
s_{\text{lat}} &= \text{usmht}(\Delta y, \phi_4, 1000, 1) + 1 \label{eq:slat} \\
s_{\text{ramp}} &= \text{usmht}\left( \frac{\Delta x_{\text{ramp}}}{v_{\text{MA}}}, \phi_5, 1000, 1 \right) + 1 \label{eq:sramp}
\end{align}

where:
\begin{itemize}[leftmargin=1.5em]
    \item \(\Delta y\): Lateral distance between MA and Lag.
    \item \(\Delta x_{\text{ramp}}\): Longitudinal distance from MA to ramp end.
    \item \(\phi_4, \phi_5\): Tunable parameters shaping scaling sensitivity.
\end{itemize}

\subsubsection{Merging Vehicle Payoffs}

The MA chooses between two actions: keeping straight or changing lanes. The corresponding payoffs are:

\begin{equation}
\begin{split}
P^*_{\text{Keep}} = 0.5 \Big( 
\text{usmht}\left( \Psi_{\text{MA,Lead}}, \phi_1, 1, 1 \right) \\
\quad + \text{usmht}\left( \Psi_{\text{MA,Lag}}, \phi_8, 1, -1 \right)
\Big)
\end{split}
\label{eq:pkeep}
\end{equation}

\begin{align}
P^*_{\text{Change}} &= -P^*_{\text{Keep}} + \frac{s_0 + v_{\text{MA}} T}{\Delta x_{\text{ramp}}} \label{eq:pchange}
\end{align}

where:
\begin{itemize}[leftmargin=1.5em]
    \item \(P^*_{\text{Keep}}\): Incentive for remaining in the ramp based on gaps to lead and lag vehicles.
    \item \(P^*_{\text{Change}}\): Incentive for merging, increasing as the ramp end approaches.
\end{itemize}

\subsubsection{Decision Window}

To prevent oscillations between decisions, we implemented a fixed decision window defined as:

\begin{equation}
\Delta t_{\text{decision}} = t_{\text{window}} + \mathcal{N}(0, \sigma^2)
\end{equation}

The time window holds the selected behavior constant for a set time (we set it to 2 seconds), optionally perturbed with Gaussian noise. This mechanism enhances decision stability and realism. By tuning a small number of payoff shaping parameters, users can easily control the aggressiveness, cooperativeness, or passivity of simulated drivers, enabling systematic testing of planning algorithms across a range of interaction styles.

\subsection{Bounded Rationality} \label{sec:rationality}
To introduce further stochastic behavior and enable change-of-mind phenomena, we incorporated bounded rationality into the model. We implemented this mechanism using a QRE framework, following the approach presented in \cite{BRGTD}:

\begin{equation}
    q^u_{\mathrm{Lag}}(f) =  \frac{e^{Q^E_{\mathrm{Lag}}(S,f)/ \beta }}{\sum_{f'\in F}e^{Q^E_{\mathrm{Lag}}(S,f')/ \beta}}
\label{eq:boundedrationality}
\end{equation}
where $q^u_{\mathrm{Lag}}(f)$ is the updated probability of Lag choosing action $f$, $Q^E_{\mathrm{Lag}}(S,f')$ is the lag's expected payoff for a given action $f'$, $\beta$ is the bounded rationality coefficient, and $S$ and $F$ are the merger's and lag's action sets, respectively. The lag’s expected payoff \( Q^E_{\mathrm{Lag}}(S,f) \) for a given action \( f \) is calculated as:

\begin{equation}
    Q^E_{\mathrm{Lag}}(S,f) = \sum_{s\in S}q_{\mathrm{Lag}}(s,f)Q_{\mathrm{Lag}}(s,f)
\label{eq:expectedpayoff}
\end{equation}

where $q_{\mathrm{Lag}}(s,f)$ and $Q_{\mathrm{Lag}}(s,f)$ are the lag's probability and payoff, respectively, for merger action $s$ and lag action $f$. The probabilities are derived by solving for the Nash equilibrium.

During simulation, low values of \( \beta \) (e.g., 0.01) correspond to highly rational behavior, where the agent almost always selects the highest payoff action. As \( \beta \) increases, decision randomness also increases, causing the action probabilities to converge toward a uniform distribution. When \( \beta \) becomes large, the actor behaves increasingly irrationally, assigning nearly equal probabilities to all options. This behavior enables modeling ``change-of-mind'' phenomena, where drivers may inconsistently switch decisions even within the same merging scenario.

\section{Model Calibration and Results} \label{sec:dataset}
\subsection{Dataset Selection}

Since we aimed to model real-world, humanlike interactions between merging vehicles and main lane vehicles, we selected a dataset that offered rich diversity in on-ramp merging behavior. 
The \textbf{HOMER} dataset~\cite{BRGTD} is a large U.S.-based on-ramp interaction dataset. It includes varying traffic densities, ramp geometries, vehicle types, and speed profiles. 
To ensure robust and meaningful calibration, we filtered the \textbf{HOMER} dataset to include only merging events that lasted at least 3 seconds. This filtering ensured that each selected lag-merger interaction contained sufficient temporal context for decision modeling. In total, from eight ramp sites, 9,051 lag actors were included in model calibration, with an average lag merging duration of 7.2 seconds.

\subsection{Ground Truth Generation}

The data mainly has timestamped state information of the traffic agents and scene. However, since we are focusing on the higher level decision-making of agents, it is important to process the data to extract the observed actions. 
We thus translated continuous vehicle trajectories into interpretable ground-truth behavior labels. We generated these labels by analyzing the time gap profile between the lag vehicle and its leader. Specifically, we defined the following behavior classes:

\begin{itemize}
    \item \textbf{Do nothing}: The lag vehicle maintained a stable time gap to the leader.
    \item \textbf{Yield behind}: The lag vehicle increased its time gap, allowing the merger to pass ahead.
    \item \textbf{Yield ahead or block}: The lag vehicle decreased its time gap, either to preempt the merger or to block it from entering.
\end{itemize}

 We applied a 2-second Savitzky-Golay filter \cite{savitzky1964smoothing} to smooth the raw time gap series. This reduced measurement noise while preserving relevant behavioral patterns. To classify behavior epochs, we applied thresholds to both the \textit{rate of change} and the \textit{total change} in the time gap. Specifically, we used a minimum average rate of $\pm 0.08$ seconds/second and a total change threshold of $\pm 1.0$ seconds to determine whether the gap was increasing, decreasing, or stable.


 Although MR-LDM treats \textit{yield ahead} and \textit{block} as distinct decisions, our segmentation process could not reliably separate them based on longitudinal data alone. Both behaviors exhibit a decreasing time gap, and without lateral or intent-specific cues, the segmentation algorithm grouped them under a single \textit{gap closing} behavior. During evaluation, we resolved this ambiguity by 
taking the highest output probability between \emph{yield ahead} and \emph{block} for each \emph{gap closing} observation.

\subsection{Calibration Procedure}

We calibrated the MR-LDM model using a bi-level optimization approach that aligns model-predicted behavior distributions with real-world observations. This framework follows methodologies proposed in prior decision modeling studies \cite{BRGTD} and \cite{chen2023}, which similarly used a bi-level optimization to estimate payoff parameters based on observed driver behaviors. 

At the \textit{lower level}, we computed the mixed Nash equilibrium at each decision point to determine the probability distribution over the possible lag behaviors. 
Given the merger’s action set and the current game state, this step produced a probabilistic response profile from the lag actor for each observation in the dataset.

At the \textit{upper level}, we optimized the model parameters $\{\phi_1, \dots, \phi_8, \tau\}$ to minimize the deviation between the predicted behavior probabilities and the observed behavior labels. We defined the objective as:

\[
\min_{\phi} \sum_{k=1}^{N} \left(1 - q_{a_k}\right)
\]

where $q_{a_k}$ denotes the probability assigned by the model to the correct action $a_k$ at observation $k$, and $N$ is the total number of observations. We solved this optimization using MATLAB’s \texttt{MultiStart} \cite{MultiStart} global search framework with the \texttt{fmincon} solver.

We performed two forms of calibration:
\begin{itemize}
    \item \textbf{Global calibration}: A single parameter set was optimized across all lag-merger interactions to represent an average driver profile.
    \item \textbf{Lag-specific calibration}: Each lag vehicle’s behavior was calibrated independently to capture driver heterogeneity.
\end{itemize}


\subsection{Results}\label{sec:results}

We evaluated the MR-LDM model's performance by comparing it directly with the baseline RGLC model~\cite{RGLC}. We assessed model accuracy using the behavior classifications obtained from the ground truth segmentation described earlier. We use the \textbf{Mean Absolute Error (MAE)} metric, which is the average deviation of model-predicted probabilities from perfect accuracy (probability = 1). Table~\ref{tab:modelComparison} summarizes the comparative performance between MR-LDM and RGLC.


\begin{table}[htbp]
    \centering
    \begin{threeparttable}
    \caption{Comparison of MR-LDM vs. RGLC}
    \label{tab:modelComparison}
    \begin{tabular*}{0.9\columnwidth}{@{\extracolsep{\fill}}lccc}
    \toprule
    \textbf{Model} & \textbf{MAE*} & \textbf{\# Params} & \textbf{Speed**}\\
    \midrule
    RGLC~\cite{RGLC} & 0.27 (0.87) & 26 & \textbf{0.3 ms}\\[0.5ex]
    MR-LDM & \textbf{0.17} & \textbf{9} & 0.5 ms\\
    \bottomrule
    \end{tabular*} 
    \begin{tablenotes}
      \small
      \item[*] The first MAE value for RGLC only includes data with behaviors considered by the model. The second value additionally includes \emph{do nothing} behavior.
      \item[**] Speed per iteration for one actor in MATLAB R2024b.
    \end{tablenotes}
    \end{threeparttable}
\end{table}

The MR-LDM achieved approximately 10\% higher accuracy compared to RGLC. This improvement highlights the advantage of explicitly modeling a wider set of driver behaviors—such as \textit{yield ahead} and \textit{block}—and using the \textit{PTH} metric within the payoff function. MR-LDM uses only nine tunable parameters compared to 26 in RGLC, enabling faster calibration and easier tuning without sacrificing performance.



\begin{figure}
 \vspace{3mm}
    \centering
    \includegraphics[width=0.49\textwidth,keepaspectratio]{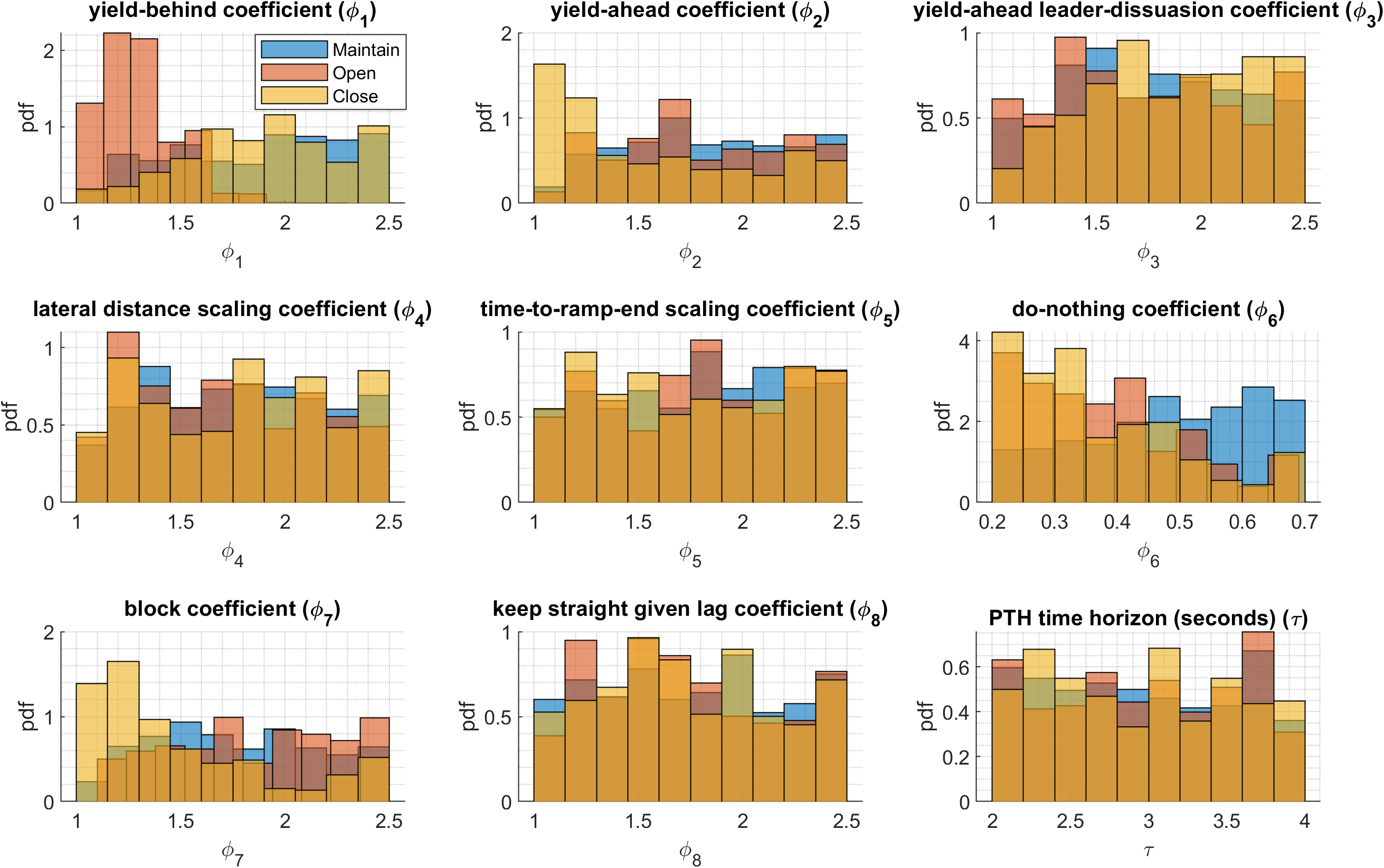}
    \caption{Parameter distributions from MR-LDM lag-specific calibration. The results have been separated based on the behavior $\mathrm{mode}$ found in the ground truth and are discussed in Section~\ref{sec:results}.}
    \label{fig:MRLDMdists}
\end{figure}

To investigate driver heterogeneity further, we analyzed parameter distributions derived from lag-specific calibration. Figure~\ref{fig:MRLDMdists} illustrates these parameter distributions grouped by dominant ground-truth behavior classes: maintaining, opening, and closing gaps. Clear distinctions emerged:
\begin{itemize}
    \item Drivers predominantly exhibiting \textit{do nothing} behavior had higher values of the do-nothing parameter $\phi_6$.
      \item Drivers frequently performing \textit{yield behind} had lower values of the yield-behind parameter $\phi_1$.
    \item Aggressive or blocking drivers exhibited lower values of the blocking parameter $\phi_7$ and yield-ahead parameter $\phi_2$, alongside higher values of the leader-dissuasion parameter $\phi_3$.
\end{itemize}

These insights validate MR-LDM's ability to represent diverse driving styles, making it a suitable tool for simulating realistic merging scenarios and testing autonomous vehicle decision-making algorithms.

\section{Behavior Execution} \label{sec:beh-ex}

The MR-LDM model is a higher-level decision model that outputs a discrete longitudinal behavior decision at each time step. To integrate these decisions into a continuous vehicle simulation environment, we developed a corresponding behavior execution strategy based on the previously proposed MR-IDM \cite{MRIDM}. 
The MR-LDM module acted as a decision layer, feeding behavior labels to MR-IDM at every step. This enabled seamless interaction between high-level game-theoretic reasoning and low-level vehicle dynamics.

\begin{figure}
\vspace{4mm}
    \centering
    \includegraphics[width=0.5\textwidth,keepaspectratio]{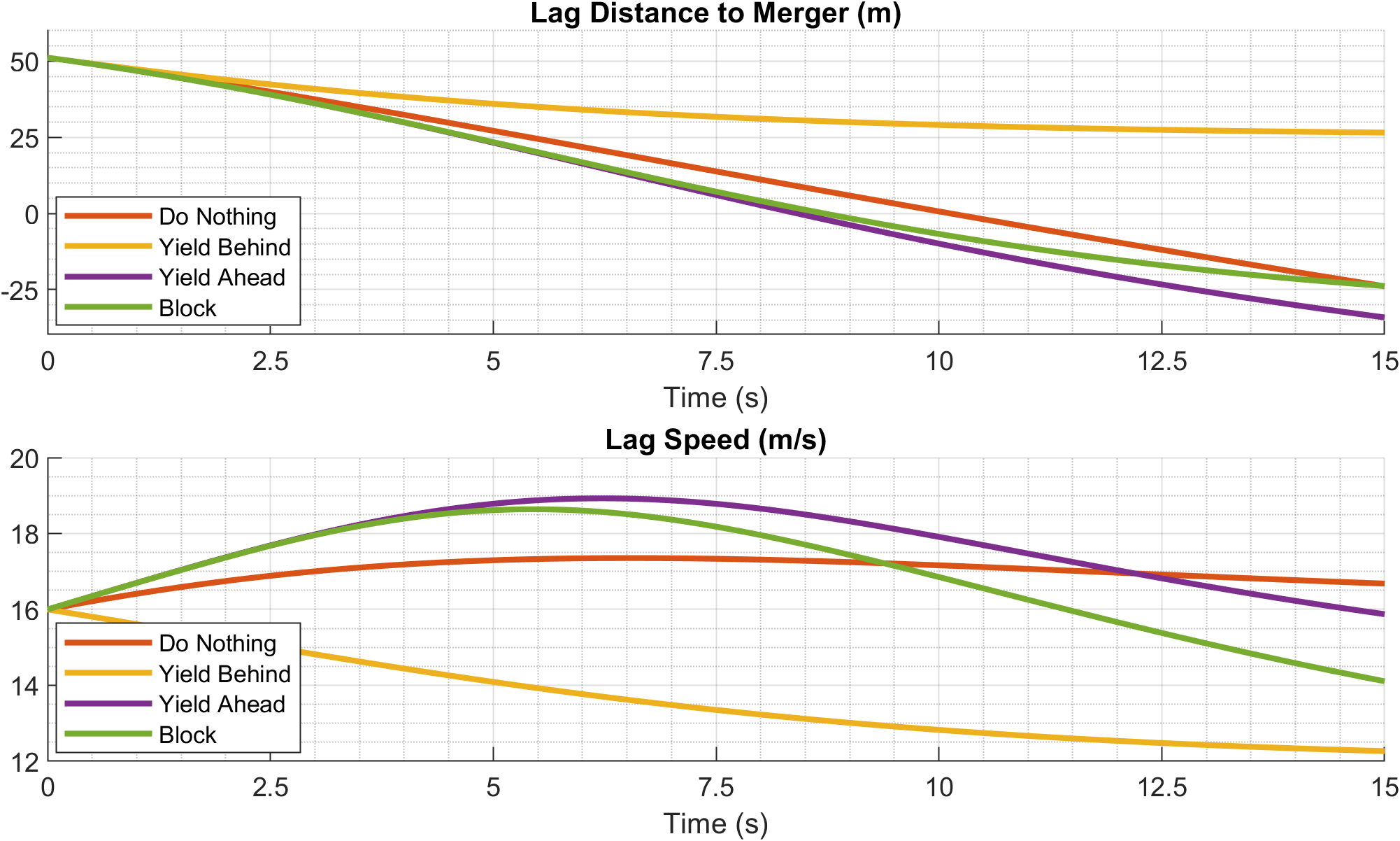}
    \caption{Lag behavior execution for four parameter sets. In this example, a lag vehicle is initialized with a merger 51 meters ahead. Four simulations are run with varying MR-LDM parameters. MR-IDM parameters are held constant.}
    \label{fig:MRLDMexample}
\end{figure}

We defined behavior-specific execution logic for all four modeled decisions:
\begin{itemize}
    \item \textbf{Yield Behind}: We directly applied the MR-IDM formulation, targeting a safe gap to the merger. This was the default behavior MR-IDM was originally designed to handle.

    \item \textbf{Yield Ahead}: We configured the lag vehicle to reduce the gap to its leader in the main traffic lane, creating space for the merger to enter behind. Specifically, we modified the MR-IDM inputs to target a new time gap that was shorter by at least half the distance between the merger and the leader. We also increased the desired speed parameter ($v_0$) by up to 5~m/s and reduced the safe time headway ($T$) and standstill distance ($s_0$) to allow the lag vehicle to move closer to its leader.

    \item \textbf{Block}: We executed this behavior similarly to yield ahead but targeted a gap equal to the distance between the merger and the leader. This adjustment positioned the lag vehicle beside the merger, effectively preventing it from changing lanes. The success of the blocking maneuver can be tightly coupled with the MR-IDM parameters. For example, in Figure \ref{fig:MRLDMexample}, the blocking behavior would maintain a longer period with a small relative distance to the merger if the magnitudes of parameters $a$ and $b$ were increased, allowing for higher acceleration and deceleration.

    \item \textbf{Do Nothing}: We applied the lag’s default MR-IDM parameters without regard to the merger. However, if a prior non-default behavior (e.g., yield or block) had modified the current spacing or speed, we preserved those values in a separate parameter set. The do-nothing behavior then maintained this adjusted state by reapplying the modified parameters rather than resetting to the original ones.
\end{itemize}
While discontinuities in acceleration can occur if the model changes decisions, the introduced decision window helps reduce this. Also, decision changes are less common for higher rationality in the model.



\begin{figure}
    \centering
    \includegraphics[width=0.5\textwidth,keepaspectratio]{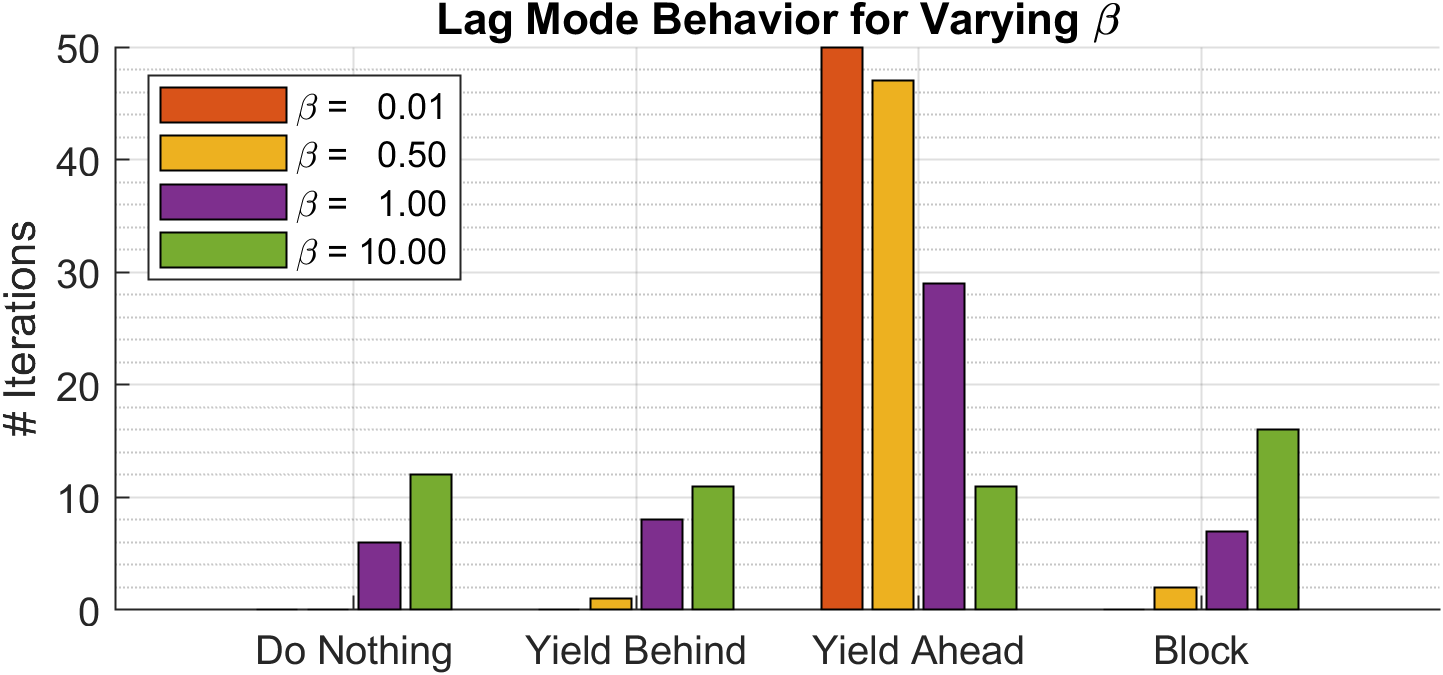}
    \caption{Using the lag example from Figure \ref{fig:MRLDMexample}, this figure shows the mode behavior chosen by the MR-LDM for different values of the bounded rationality coefficient $\beta$. The simulation was run for 50 iterations for each value of $\beta$. Note that, as $\beta$ increases, the actor's behavior becomes less deterministic.}
    \label{fig:MRLDMBoundedRationalityexample}
\end{figure}

To evaluate simulation scalability, we integrated the proposed models into a high-fidelity simulation environment capable of reproducing complex highway traffic scenarios. This setup enabled software-in-the-loop testing of autonomous vehicle (AV) control and decision-making algorithms under diverse conditions. We used IPG-CarMaker to simulate vehicle dynamics and road environments at high fidelity, while MATLAB-Simulink executed the traffic behavior models and vehicle control logic. The integrated framework successfully simulated a highway merge scenario involving 20 vehicles, each operating an independent traffic behavior model and full vehicle dynamics in real time. We ran these simulations on a standard laptop equipped with a 64-bit operating system, an Intel(R) Xeon(R) W-2123 CPU @ 3.6 GHz, and 16 GB RAM. Thanks to its lightweight decision structure and efficient payoff computation, MR-LDM supports real-time execution even when integrated with high-fidelity vehicle dynamics and multi-agent traffic simulations.

\section{Conclusion and Future Work}\label{sec:conclusion}
In this work, we introduced the MR-LDM model, a game-theoretic framework for simulating the longitudinal behavior of a main-lane lag vehicle with a merging vehicle on a highway on-ramp. MR-LDM extends prior models by explicitly modeling four discrete behaviors - \emph{yield behind}, \emph{yield ahead}, \emph{block}, and \emph{do nothing} - which cover a large set of observed behaviors whether the target traffic is polite and yielding, aggressive, or indifferent. We developed interpretable and bounded payoff functions using a modified hyperbolic tangent formulation and introduced \textit{PTH} as a robust input metric. To simulate realistic decision variability, we incorporated stochastic behavior using QRE.

We validated the model on a diverse real-world dataset and demonstrated that MR-LDM outperforms the RGLC baseline in both classification accuracy and behavioral coverage. We also calibrated MR-LDM at both global and individual vehicle levels, showing that the model captures driver heterogeneity through interpretable parameter sets. Furthermore, we integrated the model into a high-fidelity vehicle simulation environment and demonstrated its scalability in real-time, software-in-the-loop simulation involving 20 independently controlled agents.

Currently we have decoupled the longitudinal and lateral decision models \cite{BRGTD}. Some works argue they should remain this way \cite{ALI2019220}, but future work could focus on combining these into a single decision-modeling framework. We also aim to extend MR-LDM to control not only the type of decision, but also its severity, allowing finer-grained control over gap opening or blocking behavior. Additionally, we intend to explore online adaptation of driver models through real-time parameter tuning to better reflect evolving behaviors in dynamic environments. The model can potentially also be extended in its usage in an interaction-aware planning module for a merging function of an autonomous car to help intent inference and prediction of traffic agents. 

\section{Acknowledgement}\label{sec:acknowledgement}
We would like to thank Travis Terry, Pawan Kallepalli, Lauri Mustonen, Takayasu Kumano, Yosuke Sakamoto, and Yuji Yasui for their contributions to the project.

\bibliographystyle{IEEEtran.bst} 
\bibliography{reference/ref.bib}

\end{document}